\documentclass[10pt,twocolumn,letterpaper]{article}

\usepackage[pagenumbers]{cvpr} %

\newcommand{\teaminfo}[3]{
    \begin{table}[H]
        \vspace{-2mm}
        \begin{tabularx}{\linewidth}{@{}lX@{}}
            \toprule
            \textbf{Team:} & {#1} \\
            \textbf{Members:} & {#2} \\
            \textbf{Affiliation:} & {#3} \\
            \bottomrule
        \end{tabularx}
        \vspace{-2mm}
    \end{table}
    }

\usepackage{multirow}
\usepackage{colortbl}
\usepackage{xcolor}
\usepackage{pifont}
\usepackage{makecell}
\usepackage{adjustbox}
\usepackage{threeparttable}
\usepackage{setspace}
\usepackage{tabularx}
\usepackage{float}

\usepackage{pythonhighlight}
\usepackage[linesnumbered,ruled,vlined]{algorithm2e}

\makeatletter
\newcommand{\thickhline}{%
    \noalign {\ifnum 0=`}\fi \hrule height 1pt
    \futurelet \reserved@a \@xhline
}
\makeatother
\newcommand{\J}{$\mathcal{J}$\xspace}
\newcommand{\F}{$\mathcal{F}$\xspace}
\newcommand{\JF}{$\mathcal{J}\&\mathcal{F}$\xspace}

\newcommand{\JFnd}{$\mathcal{J}\&\dot{\mathcal{F}}_d$\xspace}
\newcommand{\JFnr}{$\mathcal{J}\&\dot{\mathcal{F}}_r$\xspace}
\newcommand{\Fn}{$\dot{\mathcal{F}}$\xspace}
\newcommand{\JFn}{$\mathcal{J}\&\dot{\mathcal{F}}$\xspace}

\definecolor{cvprblue}{rgb}{0.21,0.49,0.74}
\usepackage[pagebackref,breaklinks,colorlinks,allcolors=cvprblue]{hyperref}

\def\paperID{*****} %
\usepackage[misc]{ifsym}
\def\confName{CVPR}
\def\confYear{2025}
\usepackage[accsupp]{axessibility}  %
\usepackage{fontawesome5}
\usepackage{array}
\definecolor{gold}{RGB}{255, 205, 0}
\definecolor{silver}{RGB}{192, 192, 192}
\definecolor{bronze}{RGB}{205, 127, 50}  %
\title{LSVOS 2025 Challenge Report:\\ Recent Advances in Complex Video Object Segmentation
}

\author{\vspace{5pt}
    Chang Liu\textsuperscript{*},\ 
    Henghui Ding\textsuperscript{*}$^{\textrm{\Letter}}$,\ 
    Kaining Ying\textsuperscript{*},\ 
    Lingyi Hong\textsuperscript{*},\ 
    Ning Xu\textsuperscript{*},\ 
    Linjie Yang\textsuperscript{*},\ 
    Yuchen Fan\textsuperscript{*},\\
    Mingqi Gao,\ \  Jingkun Chen,\ \  Yunqi Miao,\ \  Gengshen Wu,\ \  Zhijin Qin,\ \  Jungong Han,\\
    Zhixiong Zhang,\ \  Shuangrui Ding,\ \  Xiaoyi Dong,\ \  Yuhang Zang,\ \  Yuhang Cao,\ \ Jiaqi Wang,\\
    Chang Soo Lim,\ \ Joonyoung Moon,\ \ Donghyeon Cho,\qquad Tingmin Li,\ \  Yixuan Li,\ \  Yang Yang,\\
    An Yan,\ \ Leilei Cao,\ \  Feng Lu,\ \ 
    Ran Hong,\ Youhai Jiang,\ \  Fengjie Zhu,\quad Yujie Xie,\ \  Hongyang Zhang,\\
    Zhihui Liu,\ \  Shihai Ruan,\quad Quanzhu Niu,\ \  Dengxian Gong,\ \  Shihao Chen,\ \ Tao Zhang,\ \  Yikang Zhou,\\
    Haobo Yuan,\ \  Lu Qi,\ \  Xiangtai Li,\ \  Shunping Ji,\qquad
    Ran Hong,\ \ Feng Lu,\ \  Leilei Cao,\ \  An Yan,\\  \vspace{5pt}
    Alexey Nekrasov,\ \  Ali Athar,\ \  Daan de Geus,\ \  Alexander Hermans,\ \ Bastian Leibe\\
\href{https://lsvos.github.io/}{https://lsvos.github.io/}
}

\begin{document}
\maketitle
\renewcommand{\thefootnote}{\fnsymbol{footnote}}
\footnotetext[1]{Organizers of the ICCV 2025 LSVOS Challenge. Following authors are top 3 team members of the three challenge tracks.}
\footnotetext[0]{${\textrm{\Letter}}$ Corresponding to Henghui Ding (henghui.ding@gmail.com), the Institute of Big Data, Fudan University, Shanghai, China.}

\begin{abstract}
This report presents an overview of the 7th Large-scale Video Object Segmentation (LSVOS) Challenge held in conjunction with ICCV 2025. Besides the two traditional tracks of LSVOS that jointly target robustness in realistic video scenarios: Classic VOS (VOS), and Referring VOS (RVOS), the 2025 edition features a newly introduced track, Complex VOS (MOSEv2). Building upon prior insights, MOSEv2 substantially increases difficulty, introducing more challenging but realistic scenarios including denser small objects, frequent disappear/reappear events, severe occlusions, adverse weather and lighting, etc., pushing long-term consistency and generalization beyond curated benchmarks. The challenge retains standard $\mathcal{J}$, F, and $\mathcal{J\&F}$ metrics for VOS and RVOS, while MOSEv2 adopts $\mathcal{J\&\dot{F}}$ (the average of region similarity $\mathcal{J}$ and adaptive boundary accuracy $\mathcal{\dot{F}}$) as the primary ranking metric to better evaluate objects across scales and disappearance cases. We summarize datasets and protocols, highlight top-performing solutions, and distill emerging trends, such as the growing role of LLM/MLLM components and memory-aware propagation, aiming to chart future directions for resilient, language-aware video segmentation in the wild. 

\end{abstract}
    
\section{Introduction}
\label{sec:intro}

Video object segmentation (VOS) ~\cite{MOSE,MeViS,li2024transformer,wu2024towards,ravi2024sam,hesham2025exploiting} targets identifying and following an user-selected objects throughout a video sequence. Despite impressive results on classic benchmarks such as DAVIS and YouTube-VOS, previous research feature large, salient, or relatively isolated targets, leaving open questions about robustness in unconstrained, real-world footage. The Large-scale Video Object Segmentation (LSVOS) challenge series was created to address this gap by providing a community venue that refreshes tasks and datasets in step with emerging needs, enabling rigorous comparisons and spotlighting unsolved problems. 

To explicitly stress real-world complexity, in last year's challenge, we have introduced the MOSE dataset (MOSEv1)~\cite{MOSE} that contains crowded scenes, frequent occlusions, and object disappearance/reappearance, revealing that many competitive systems still struggle once video content departs from canonical, single-object, or low-clutter settings. Additionally, the LVOS dataset~\cite{hong2023lvos} is also employed by LSVOS to evaluate and enhance model's performance on longer videos. This line of work reframed VOS evaluation toward harder cases and motivated new algorithmic designs emphasizing robustness and long-term consistency. 
Building on these insights, in this year's 7th LSVOS Challenge, we specially highlights \textbf{MOSEv2}, a substantially more challenging dataset designed to advance VOS under real-world conditions. As a successor to MOSEv1, this new benchmark is designed to be significantly more difficult, incorporating a wide range of previously under-represented challenges, such as adverse weather, poor lighting, \textit{etc.}. For more accurate assessment, MOSEv2 also introduces new metrics that better capture performance across object scales and disappearance cases. These additions aim to push VOS research toward models that transfer robustly beyond curated benchmarks.

Along with MOSEv2, recent breakthroughs in large language models (LLMs) and their multimodal counterparts are increasingly steering the evolution of computer vision~\cite{shuai2024survey}. In parallel, foundation-style vision models such as SAM2~\cite{ravi2024sam} capitalize on huge training data to deliver strong out-of-distribution generalization. Building on this momentum, our challenge additionally emphasizes Referring Video Object Segmentation (RVOS)~\cite{MeViS}, aim to evaluate language-grounded, interactive segmentation on MeViS. Together, these tracks advance toward more resilient and unified vision systems that integrate perception, interaction, and linguistic grounding. With the LSVOS's most challenging and realistic dataset to date, we aims to further push the boundaries of VOS research.

This year's LSVOS Challenge offers three tracks: newly introduced Complex VOS (MOSEv2) track, and our traditional tracks, Classic VOS (VOS) and Referring VOS (RVOS), to assess generalization across both established setups and the newly emphasized real-world scenarios. This report presents the datasets and protocols, summarizes the top solutions, and distills lessons and open challenges revealed by the 2025 edition, with the goal of VOS research toward models that transfer robustly beyond curated benchmarks.

\section{The LSVOS 2025 Challenge}

\subsection{Challenge Tracks}

\indent\textbf{Track 1: The MOSEv2 Track}
\vspace{2mm}

\noindent\textbf{\textit{MOSEv2}}~\cite{ding2025mosev2} is the successor of the MOSEv1 dataset, with enhanced scale, diversity, and complexity. MOSEv2 contains 5,024 videos, 701,976 high quality masks, and 10,074 objects spanning 200 categories. Beyond scale, it is constructed to stress failure modes that standard benchmarks seldom capture: smaller and denser targets that challenge spatial resolution; frequent disappear/reappear events that test long term temporal reasoning; severe occlusions and crowding that confound association; and adverse conditions, including rain, snow, fog, low-light/nighttime, and even underwater footage. It further broadens difficulty with multi shot sequences that require cross-cut consistency, camouflaged objects that reduce figure, ground contrast, non-physical targets (\textit{e.g.}, shadows and reflections), and scenes that demand external knowledge (\textit{e.g.}, understanding function or context) to disambiguate candidates.
For the challenge track, MOSEv2 is used to probe robust generalization across diverse real-world scenarios and usage paradigms. Systems must sustain accurate masks over long horizons, handle multi-object interactions without identity swaps, and maintain stability when appearance, illumination, or viewpoint shifts abruptly. Evaluations cover five settings, enabling a broad look at model behavior under different supervision and interaction regimes. 

\vspace{2mm}
\noindent\textbf{Track 2: VOS Track}
\vspace{2mm}

\noindent\textbf{\textit{Complex Video Object Segmentation (MOSE)}}~\cite{MOSE} aims to track and segment objects in videos of complex environments. This track is based on the MOSE~\cite{MOSE} dataset, which is a new video object segmentation benchmark designed to study object tracking and segmentation in complex, real-world scenes. Unlike previous video segmentation datasets~\cite{Pont-Tuset_arXiv_2017,youtube_vos} that focus on salient and isolated objects, MOSE features crowded environments, frequent occlusions, and object disappearances. It consists of 2,149 video clips and 5,200 objects across 36 categories, with over 430,000 high-quality segmentation masks. MOSE challenges existing VOS models and highlights the performance gap in complex scenarios, encouraging further research into robust segmentation techniques. This year's testing set is a part of MOSE testing set, but with more challenging newly taken data added. The ground truths of all videos in the testing sets are confidential and has never been released before. 

\vspace{2mm}
\noindent\textbf{Track 2: MeViS Track}
\vspace{2mm}

\noindent\textit{\textbf{Motion Expression guided Video Segmentation (MeViS)}} \cite{MeViS,ding2025mevis} focuses on segmenting objects in video based on a sentence describing the motion of the objects, which is based on the MeViS dataset. The MeViS dataset~\cite{MeViS,ding2025mevis} is a large-scale benchmark designed for motion-guided language-based video object segmentation. Unlike previous referring image segmentation or referring video segmentation works~\cite{GRES,VLTPAMI,ding2021vision,ISFP,wang2025hierarchical,wu2024towardstip,M3Att,Zhang_2021_ICCV,he2024segpoint,he2023grec,liu2024primitivenet,liu2024referring,he2024refmask3d,3DGRES} that focus on static object attributes, MeViS emphasizes motion-centric language expressions to identify and segment target objects in complex video scenes. It features a wide range of motion expressions paired with videos containing crowded and dynamic environments. Benchmarking results show that existing referring video object segmentation methods struggle with this task, highlighting the need for new methods that can better leverage motion as a primary cue in language-guided video segmentation. Similarly, the testing set of this track comes from MeViS testing set, with newly added videos and confidential ground-truths. 

\subsection{Evaluation}
Two traditional tracks, VOS and RVOS, are evaluated using standard metrics consistent with prior LSVOS challenges~\cite{ding2024pvuw,ding2024lsvos} and benchmarks such as DAVIS~\cite{Pont-Tuset_arXiv_2017} and YouTube-VOS~\cite{youtube_vos}. Specifically, we adopt region similarity ($\mathcal{J}$), contour accuracy ($\mathcal{F}$), and their average ($\mathcal{J}\&\mathcal{F}$), with $\mathcal{J}\&\mathcal{F}$ serving as the primary ranking metric. All evaluations of the two tracks are conducted on the publicly accessible CodaLab~\cite{codalab_competitions_JMLR} platform. 

Notably, the MOSEv2 dataset introduces 5 new metrics, including adaptive boundary accuracy ($\dot{\mathcal{F}}$) that improves the original boundary accuracy for objects of different sizes, and metrics designed for disappear/reappear cases. In the LSVOS challenge, while we still report the classic metrics $\mathcal{J}\&\mathcal{F}$ for methods in this track, following the validation set of MOSEv2, we use the average of the region similarity ($\mathcal{J}$) and the adaptive boundary accuracy ($\dot{\mathcal{F}}$), denoted as $\mathcal{J}\&\dot{\mathcal{F}}$, as the primary ranking metric. Please refer to the MOSEv2 paper~\cite{ding2025mosev2} for more details about these new metrics. MOSEv2 track is hosted publicly on the CodaBench~\cite{codabench} platform.

\vspace{2mm}

\section{MOSEv2 Track Top Solution}
\label{sec:mosev2_method}
The top three teams of the MOSEv2 track are reported in Table~\ref{tab:results_mosev2}. The first place team achieved a $\mathcal{J}\&\mathcal{\dot{F}}$ score of 39.89\% on the testing set.

\begin{table}[!h]
    \renewcommand\arraystretch{1.1}
    \centering
    \setlength\tabcolsep{5pt}
    \caption{Top 3 winners of the MOSEv2 Track.}
    \vspace{-3mm}
    \small
    {\begin{tabular}{rlccccc}
            \toprule
             Rank & Team  & {$\mathcal{J}$} & {$\mathcal{F}$} & {$\mathcal{J\&F}$} & {$\mathcal{\dot{F}}$} & {$\mathcal{J\&\dot{F}}$} \\
             \midrule
             \textcolor{gold}{\faTrophy} 1 & \textbf{DSS-Track} & \textbf{39.02} & \textbf{42.35} & \textbf{40.68} & \textbf{40.76} & \textbf{39.89} \\
             \textcolor{silver}{\faMedal} 2 & IXC-Seg    & 38.87 & 42.09 & 40.48 & 40.53 & 39.70 \\
             \textcolor{bronze}{\faMedal} 3 & hyu\_cvlab & 36.99 & 40.06 & 38.52 & 38.75 & 37.87 \\
            \bottomrule
        \end{tabular}}%
    \label{tab:results_mosev2}%
\end{table}%

\subsection{1st Team in MOSEv2 Track}

\teaminfo{DSS-Track}{Mingqi Gao$^{1}$, Jingkun Chen$^{2}$, Yunqi Miao$^{3}$, Gengshen Wu$^{4}$, Zhijin Qin$^{1}$, Jungong Han$^{1}$}{{$^{1}$Tsinghua University}\quad{$^{2}$University of Oxford}\quad{$^{3}$University of Warwick}\quad{$^{4}$City University of Macau}}

Our solution uses SeC~\cite{zhang2025sec}, a SAM-2-based framework with enhanced concept modelling. SeC is built based on InternVL-2.5-4B~\cite{chen2024expanding}
and SAM-2-Large~\cite{ravi2024sam2}. The former builds semantic-level memory for the targets, while the latter focuses on fine-grained cross-frame matching and strong object perception. Given an input video with $T$ frames ($\mathcal{V}=\{v_t\in \mathbb{R}^{H\times W\times 3}\}^{T}_{t=1}$) and the first-frame annotation $m_1$, SeC generates pixel-level masks of the target object for the remaining frames $\mathcal{M}=\{m_t\in \mathbb{R}^{H\times W}\}_{t=2}^T$. The inference of each frame selectively goes through two types of memory:

\vspace{-1em}
\paragraph{Long-term Grounding Memory.}
Following SAM-2, it has two parts. For example, at frame $t$, the memory includes: 1) Pixel Memory: $f_\mathrm{pm}\in \mathbb{R}^{N_l\times C\times h\times w}$, built from the first frame and frames from ($t-N_l+1$) to ($t-1$). $C$, $h$, $w$ denote the number of channels and spatial size ($h=H/16$, $w=W/16$). $N_l$ is the memory size. It encodes both dense pixel features and the predicted masks of those frames. 2) Object Memory: $f_\mathrm{om}\in \mathbb{R}^{N_l\times C}$, also built from the first frame and frames from ($t-N_l+1$) to ($t-1$). It uses intermediate features from the mask prediction process, which implicitly represent object-level information. The Pixel Memory is first flattened, then concatenated with the Object Memory to form the final memory. Given the features of the $t^\mathrm{th}$ frame $f_t\in \mathbb{R}^{C\times h\times w}$, it is enhanced with the memory via 4 layers of self-attention and cross-attention, achieving $f_t^\mathrm{enh,g}=\mathrm{Cross\_Attn}(q=\mathrm{Self\_Attn}(f_t),kv=[f_\mathrm{pm},f_\mathrm{om}])$. These features are then decoded for the segmentation result. 

Unlike SAM-2, SeC allows a much larger memory size ($N_l=22$), with 24 frames during training. The memory attention modules are adjusted accordingly. Compared to SAM-2 (memory size is 7 during training/inference), SeC is clearly better at capturing long-term cross-frame relations, which helps in complex spatiotemporal scenarios.

\vspace{-1em}
\paragraph{Concept-aware Memory.}
It is built from a set of memory frames with masks. Unlike long-term memory, it keeps at most $N_c=7$ video frames, maintained in a FIFO manner.

This step is not always active. It only runs when a major scene change is detected. For detecting changes, SeC uses the Bhattacharyya distance between HSV histograms. In the challenge, we use a threshold of 0.35 to decide whether to apply the concept-aware memory. The features for mask decoding become the mean of $f_t^\mathrm{enh,c}$ and $f_t^\mathrm{enh,g}$ when the concept memory is activated.

\subsection{2nd Team in MOSEv2 Track}

\begin{table*}[t]
\centering
\caption{The performance comparison of various baselines and SeC on the MOSEv2 validation set.
}
\label{tab:mosev2_val} %
\vspace{-2.16mm}
\begin{tabular}{l|ccccccc}
\thickhline \\[-9pt]
\textbf{Method} & \JFn & \J & \Fn & \JFnd & \JFnr & \F & \JF \\[1pt]
\hline
SAM2-L~(ZS) & 49.5 & 47.7 & 51.3 & 62.9 & 27.3 & 53.6 & 50.7 \\
SAM2-L & 49.7 & 47.9 & 51.5 & 64.5 & 27.1 & 53.8 & 50.9 \\
SAMURAI-L & 51.1 & 49.0 & 53.2 & 52.4 & 34.9 & 55.8 & 52.4 \\
DAM4SAM-L & 51.2 & 49.2 & 53.2 & 57.2 & 34.2 & 55.6 & 52.4 \\
SAM2Long-L & 51.5 & 49.6 & 53.4 & 62.5 & 30.6 & 55.8 & 52.7 \\
\textbf{SeC~(ZS)} & \textbf{53.8} & \textbf{51.9} & \textbf{55.7} & \textbf{70.4} & \textbf{34.1} & \textbf{58.4} & \textbf{55.2}\\
\thickhline
\end{tabular}
\vspace{-1mm}
\end{table*}

\teaminfo{IXC-Seg}{Zhixiong Zhang$^{1}$, Shuangrui Ding$^{2}$, Xiaoyi Dong$^{2}$, Yuhang Zang$^{3}$, Yuhang Cao$^{3}$, Jiaqi Wang$^{3}$}{{$^{1}$Shanghai Jiao Tong University}~{$^{2}$The Chinese University of Hong Kong}~{$^{3}$Shanghai Artificial Intelligence Laboratory}}

\textbf{Method.} We employed the Segment Concept (SeC) framework~\cite{zhang2025sec}~\footnote{Since the actual used method is still under preparation for a conference paper, details are skipped for confidential.} SeC implicitly constructs a target concept from previous keyframes and integrates the conceptual reasoning capabilities of LVLMs with fine-grained pixel matching through a scene-adaptive activation strategy, which enables robust and efficient performance across complex scenarios.

\textbf{Dataset.} The dataset for this track is the challenging MOSEv2 dataset, which is designed to benchmark and advance VOS methods under realistic and complex conditions. MOSEv2 consists of 5,024 videos and over 701,976 high-quality masks for 10,074 objects across 200 categories. While amplifying the core challenges of its predecessor, such as frequent object disappearance-reappearance and severe occlusions, MOSEv2 introduces a range of new complexities. These include adverse weather (\textit{e.g.}, rain, snow), low-light scenes, multi-shot sequences, camouflaged objects, and non-physical targets like shadows and reflections. The dataset is officially partitioned into 3,666 training, 433 validation, and 614 testing videos. Our evaluation is primarily conducted on the official validation set and a designated 100-video subset of the test set.

\textbf{Main Results} As detailed in Table~\ref{tab:mosev2_val}, the SeC framework achieves leading performance on the MOSEv2 validation set, outperforming previous strong baselines, including several fine-tuned SAM2-L variants. In particular, SeC attains a primary score of 53.8 in \JFn, surpassing the SAM2Long-L by 2.3 points.

\subsection{3rd Team in MOSEv2 Track}

\teaminfo{hyu\_cvlab}{Chang Soo Lim, Joonyoung Moon, Donghyeon Cho}{{Hanyang University}}

\begin{figure*}[!t]
  \centering
  \includegraphics[width=\textwidth]{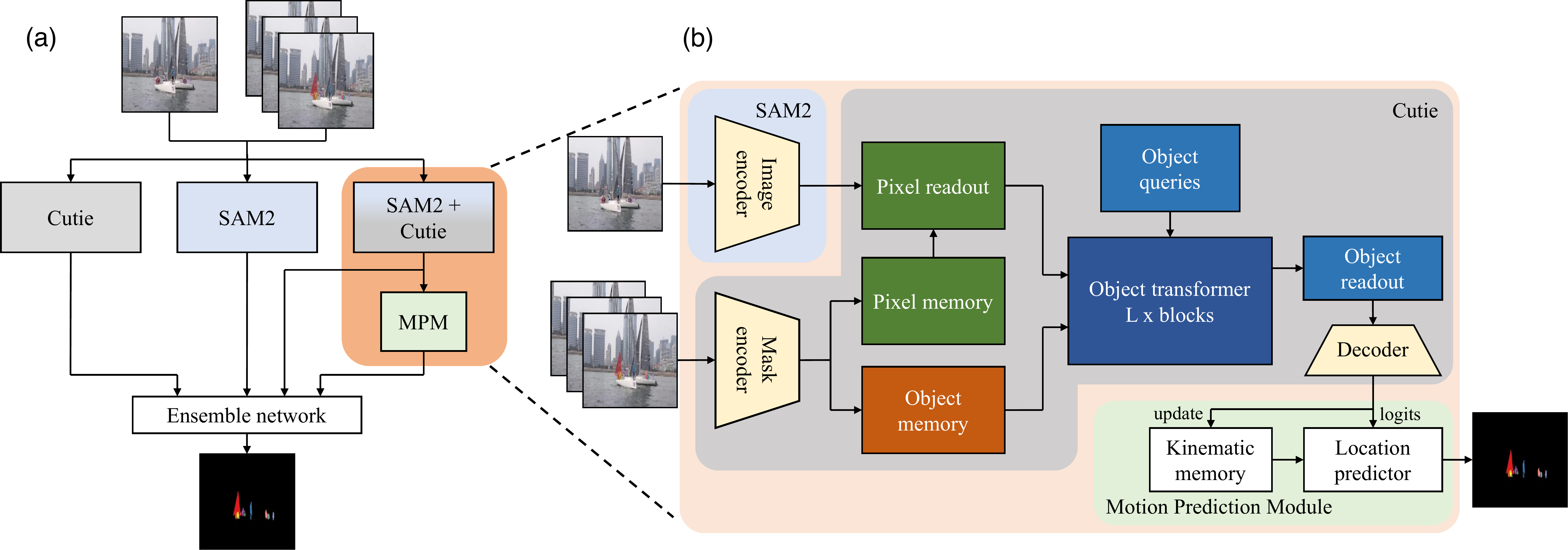}
  \caption{\textbf{The overall framework of SCOPE.} The left figure (a) illustrates our overall ensemble pipeline, while the right figure (b) shows the fusion network of SAM2 and Cutie with the proposed Motion Prediction Module (MPM).}
  \label{fig:combined}
\end{figure*}

As illustrated in Fig.~\ref{fig:combined}-(b), SCOPE is built on Cutie, where the original query encoder is replaced with the SAM2 image encoder to enrich semantic features. 
We introduce an MPM that estimates the position of objects under occlusion to improve temporal consistency. 
Finally, as shown in Fig.~\ref{fig:combined}-(a), we design an ensemble strategy that integrates Cutie, SAM2, and our variant for further performance gains.

\textbf{Enriching Feature Representation Using SAM2}
\label{sec:SAM2}
Cutie, as mentioned above, leverages object vectors that enable consistent object tracking. 
However, due to its lightweight ResNet-based image encoder, the model struggles to capture rich representations, leading to degraded segmentation performance in long-term or complex videos. 
To enrich the feature representation in Cutie, we replaced its ResNet-based encoder with the MAE pre-trained Hiera image encoder from SAM2, which provides semantically rich and robust features. 
However, the Cutie encoder and the SAM2 image encoder produce different representations in both size and distribution, requiring semantic and dimensional alignment. 
To address this, we employ a $1{\times}1$ convolutional projection layer, through which the expressive image features of SAM2 can be effectively aligned and integrated into the tracking-oriented architecture of Cutie.
\textbf{Motion Prediction Module}
\label{sec:MPM}
The model integrates the Cutie encoder and the SAM2 encoder, performs well in standard tracking cases. 
However, on challenging datasets such as MOSEv2, it often struggles when the target object temporarily disappears due to occlusion or leaving the field of view, or when multiple visually similar instances co-occur. 
To address these issues, we introduce an MPM that maintains an object-specific kinematic state (location, size, and velocity) of the target from recent frames and predicts the object position in the current frame under occlusion.
Based on this prediction, the MPM generates a Gaussian map centered at the predicted object position, which serves as a spatial prior for tracking. 
This map is combined with the segmentation logits of the VOS model via a weighted sum, guiding the model to focus on the most plausible region. 
By injecting the Gaussian map as a location-aware prior, MPM improves robustness to short-term disappearances and reduces confusion among similar objects, while remaining lightweight and optional when the prediction confidence is low.

To this end, we continuously estimate the location, size, and velocity of each target object.
For initialization, given the binary mask $M_l \in \{0,1\}^{H \times W}$ of object $l$ in the first frame, the centroid $(\tilde{x}_0^l, \tilde{y}_0^l)$ and size $(\tilde{w}_0^l, \tilde{h}_0^l)$ are computed in pixels, normalized by the image resolution $(H,W)$ to form the relative state vectors as follows:
\begin{equation}
\mathbf{x}_0^l = \left(\frac{\tilde{x}_0^l}{W}, \frac{\tilde{y}_0^l}{H}\right), 
\qquad
\mathbf{u}_0^l = \left(\frac{\tilde{w}_0^l}{W}, \frac{\tilde{h}_0^l}{H}\right).
\label{eq:normalized_state}
\end{equation}
Also, the velocity $\mathbf{v}_0^l$ is set to zero vector.

At frame $t>0$, given the predicted mask $\hat{M}_t^l$ from the VOS model, the centroid and size are similarly computed and normalized to obtain $\hat{\mathbf{x}}_t^l$ and $\hat{\mathbf{u}}_t^l$. 
The state is then updated with an exponential moving average (EMA):
\begin{equation}
\mathbf{x}_t^l = \alpha \mathbf{x}_{t-1}^l + (1-\alpha)\hat{\mathbf{x}}_t^l,
\
\
\mathbf{u}_t^l = \alpha \mathbf{u}_{t-1}^l + (1-\alpha)\hat{\mathbf{u}}_t^l,
\label{eq:update_ema}
\end{equation}
where $\alpha\in(0,1)$ balances stability and responsiveness.
The velocity is defined as the displacement of consecutive centroids:
\begin{equation}
\mathbf{v}_t^l = \mathbf{x}_t^l - \mathbf{x}_{t-1}^l
\label{eq:update_velocity}
\end{equation}
without EMA to preserve sensitivity to sudden motion.
If no valid mask $\hat{M}_t^l$ is available, the location is extrapolated using the last known velocity while the object size remains unchanged.
Finally, to incorporate the estimated kinematics, we generate a Gaussian map over the image at each frame. 
For each pixel $(i,j)$, its value is defined as
\begin{equation}
G_t^l(i,j) = \exp\left(
-\frac{(i/W - x_t^l)^2}{2\sigma_x^2}
-\frac{(j/H - y_t^l)^2}{2\sigma_y^2}
\right),
\label{eq:gaussian_map}
\end{equation}
where the center $(x_t^l,y_t^l)$ corresponds to the predicted object location and the variances $\sigma_x, \sigma_y$ are set proportional to the estimated width $w_t^l$ and height $h_t^l$.
This design adaptively scales the Gaussian distribution with the object size, yielding sharper priors for small objects and broader ones for large objects.
The Gaussian map is then integrated with the output of the segmentation network to bias the model toward the predicted region.
Specifically, let $\hat{Z}_t^l$ denote the raw logits of object $l$ in frame $t$.
We combine them with the Gaussian map through a weighted sum:
\begin{equation}
{Z}_t^l = \hat{Z}_t^l + \beta\cdot\log(G_t^l + \epsilon),
\label{eq:gaussian_weighted_sum}
\end{equation}
where $\beta$ controls the influence of the prior and $\epsilon$ is a small constant for numerical stability.
This formulation effectively increases the confidence of pixels near the predicted location while suppressing unlikely regions.
By applying this fusion to every frame, the module consistently injects location-aware information into the segmentation process.
As a result, the model can recover more gracefully from short-term disappearance (e.g., due to occlusion) and is less prone to confusion when multiple visually similar objects co-occur.
Importantly, MPM remains lightweight and optional: when the base network already produces confident predictions, the Gaussian prior has little influence, while in ambiguous cases it provides additional guidance to resolve uncertainty.

\textbf{Ensemble Network}
\label{sec:Ensemble}
To further improve robustness and accuracy, we adopt an ensemble strategy that combines the complementary strengths of multiple models. 
Specifically, as shown in Fig.~\ref{fig:combined}-(a), we integrate four components: the original SAM2, the original Cutie, SAM2 + Cutie with MPM, and SAM2 + Cutie without MPM.
The MPM-off variant is included to preserve fine-grained details, as the Gaussian map, although beneficial under occlusion, tends to oversmooth boundaries.
By combining all four models, the ensemble can retain the complementary advantages of each while reducing the impact of their individual weaknesses.

Formally, let $Z_C, Z_S, Z_{M-}, Z_{M+}$ denote the logits from Cutie, SAM2, SAM2 + Cutie without MPM, and SAM2 + Cutie with MPM, respectively, all aligned to the same spatial resolution $(H,W)$. 
These outputs are then fed into a shallow fusion module $f_\theta$:
\begin{equation}
F = f_{\theta}(Z_C, Z_S, Z_{M-}, Z_{M+})
   \in \mathbb{R}^{(N+1)\times H \times W},
\label{eq:ensemble_fusion}
\end{equation}
where $N$ is the number of object classes.
Note that~\eqref{eq:ensemble_fusion} computes a weighted combination and produces the final ensemble logits.
This design enables the ensemble to leverage the complementary strengths of all components while mitigating their individual weaknesses.

\section{VOS Track Top Solution}
\label{sec:mose_method}
The top three teams of the VOS track are reported in Table~\ref{tab:results_mose}. The first place team achieved a $\mathcal{J}\&\mathcal{{F}}$ score of 86.37\% on the testing set.

\begin{table}[!h]
    \renewcommand\arraystretch{1.1}
    \centering
    \setlength\tabcolsep{10pt}
    \caption{Top 3 winners of the VOS Track.}
    \vspace{-3mm}
    \small
    {\begin{tabular}{rlccc}
            \toprule
             Rank & Team  & {$\mathcal{J}$} & {$\mathcal{{F}}$} & {$\mathcal{J\&{F}}$} \\
             \midrule
             \textcolor{gold}{\faTrophy} 1 & \textbf{NJUST-KMG}   & \textbf{84.10} & \textbf{86.64} & \textbf{86.37} \\
             \textcolor{silver}{\faMedal} 2 & Transsion    & 83.72 & 88.59 & 86.16 \\
             \textcolor{bronze}{\faMedal} 3 & TS\_Video & 83.57 & 88.10 & 85.84 \\
            \bottomrule
        \end{tabular}}%
    \label{tab:results_mose}%
\end{table}%

\subsection{1st Team in VOS Track}

\teaminfo{NJUST-KMG}{Tingmin Li, Yixuan Li, Yang Yang}{{Nanjing University of Science and Technology}}

\begin{figure*}[t]
\centering \includegraphics[width=0.9\textwidth]{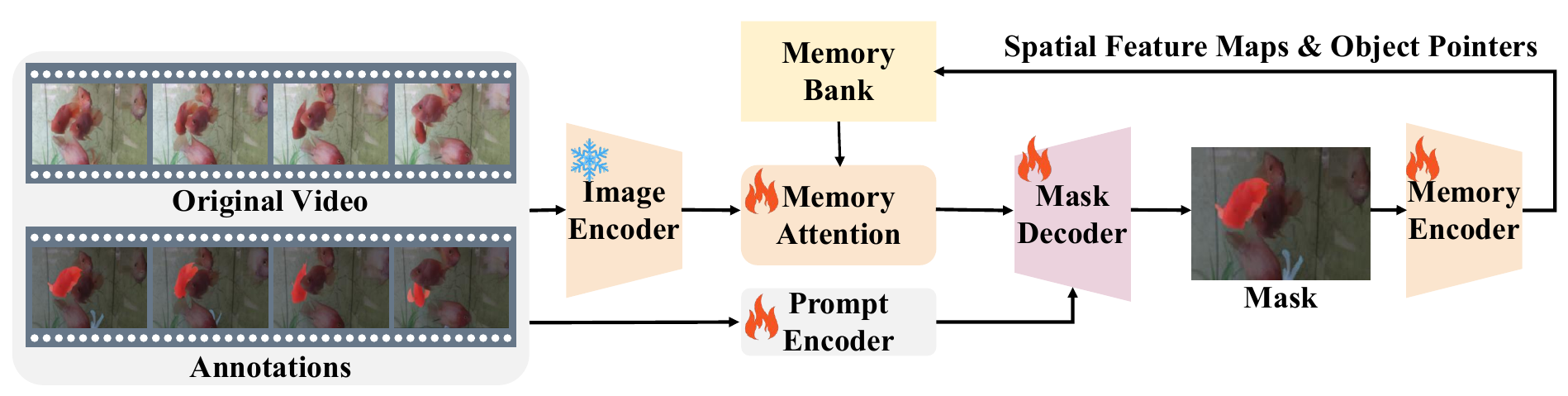} \caption{Framework of our method for the training stage.} \label{m1:fig:training} \end{figure*}

\begin{figure*}[t]
\centering \includegraphics[width=0.9\textwidth]{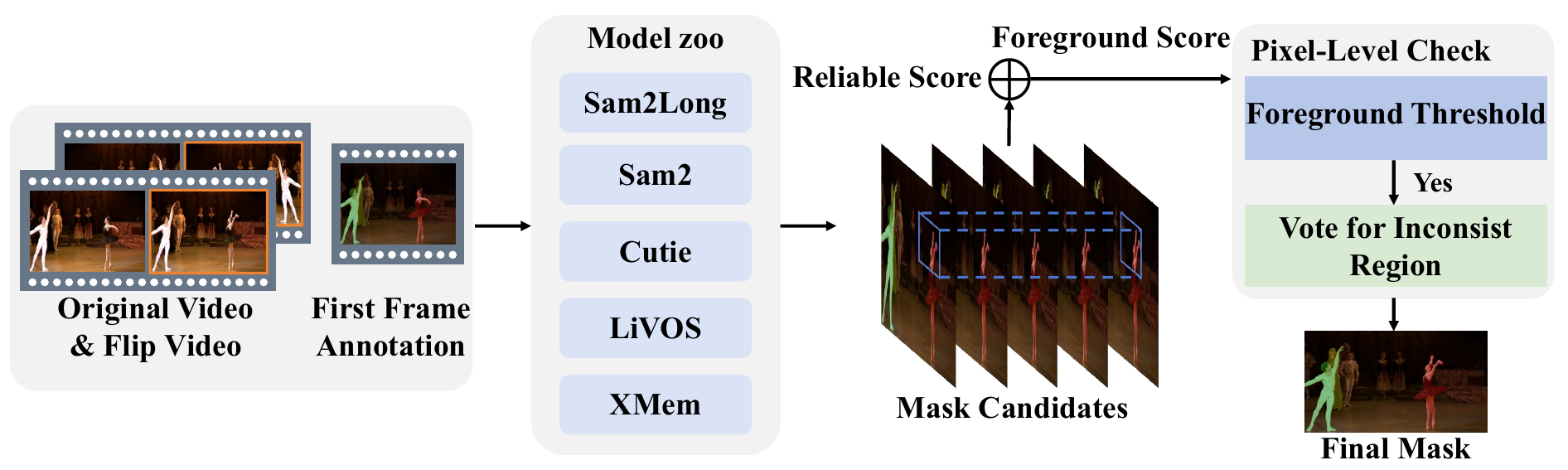} \caption{Framework of our method for the inference stage.} \label{m1:fig:test} \end{figure*}

Our training framework is illustrated in Fig. \ref{m1:fig:training}. Considering the specific characteristics of the MOSE dataset, such as frequent object disappearance and reappearance, heavy occlusions, and the presence of small and visually similar objects, we fine-tune the SAM2 model on the MOSE training set to capture these complex patterns. During inference, we introduce a confidence-guided fusion segmentation strategy, which first employs a pixel-level check mechanism to identify confident foreground pixels and then applies a voting mechanism to resolve regions with inconsistent predictions for different object IDs, thereby generating reliable results across video frames, as illustrated in the framework of the inference stage in Fig. \ref{m1:fig:test}. The detailed procedure of each stage is elaborated in the following sections.

\textbf{Training.}
As illustrated in Fig. \ref{m1:fig:training}, we adopt SAM2, a strong baseline model in the video object segmentation, serves as the foundation of our approach. The core mechanism of SAM2 lies in its memory attention module, which facilitates efficient cross-frame attention interactions and improves performance in object tracking and segmentation. Specifically, SAM2 first employs a MAE pretrained Hiera image encoder to extract rich frame-level feature representations. These frame embeddings are subsequently combined with historical frame features and object pointers to compute cross-attention, producing temporally consistent frame representations. The resulting features are then passed through a decoder to generate  segmentation masks. In parallel, the memory encoder further encodes and stores frame features, providing effective contextual information to guide accurate segmentation in subsequent frames.

Our training strategy is structured as follows. To maintain the generalization capability of SAM2 and mitigate the risk of overfitting, the image encoder is frozen, while the remaining components of the model are fine-tuned on the MOSE dataset. The model is initialized from the large checkpoint of SAM2 version 2.1 and fine-tuned for 40 epochs. During training, we employ a diverse set of data augmentation techniques, including RandomHorizontalFlip, RandomAffine, RandomResize, ColorJitter, and RandomGrayscale, to simulate the complex variations encountered in real-world scenarios, such as changes in motion patterns, occlusions, and lighting conditions. These augmentations enhance the model's robustness and its ability to generalize to challenging video sequences.

To enhance segmentation performance under complex scenarios, we adopt a multi-task loss function that simultaneously supervises both pixel-level and frame-level objectives. 
Specifically, the pixel-level supervision comprises three loss terms: Focal Loss, which identify the foreground-background pixels and adaptively emphasizes pixels that are difficult to classify, improving the model's ability to handle challenging regions; 
Dice Loss measures the region overlap between predicted masks and ground-truth annotations, which is sensitive to small object regions; and IoU Loss, which assesses the overall consistency between predicted and ground-truth masks, emphasizing holistic segmentation quality. In addition, a frame-level Classification Loss is incorporated to predict the presence or absence of target objects in each frame, providing global guidance. The combined loss function is defined as:

\begin{equation}
\mathcal{L}_{total} = \lambda_1 \mathcal{L}_{focal} + \lambda_2 \mathcal{L}_{dice} + \lambda_3 \mathcal{L}_{iou} + \lambda_4 \mathcal{L}_{cls},
\end{equation}
where $\lambda_1, \lambda_2, \lambda_3, \lambda_4$ are weighting coefficients that balance the contributions of each loss term. This multi-task optimization objectives encourages the model to capture both fine-grained pixel-level details and high-level frame-wise object presence, thereby enhancing segmentation accuracy in complex video sequences. 

\textbf{Inference.}
Observing that different VOS models exhibit complementary strengths in handling various challenges, such as occlusions, small or visually similar objects, and long-term reappearances, we propose a confidence-guided multi-model ensemble strategy to leverage their individual advantages and enhance segmentation robustness. During inference, this strategy is executed in two main phases: single-model inference and multi-model fusion.

\textbf{Phase 1: Single-Model Inference.} For initial inference, we employ five models, SAM2Long, SAM2, Cutie, LiVOS, and XMem, leveraging their complementary strengths to enhance segmentation robustness. Sam2long explicitly addresses segmentation uncertainty through a Constrained Tree Search mechanism, which selects the globally optimal segmentation path across multiple candidates for the entire video. In our experiments, the parameters are set as num\_pathway to 3, iou\_thre to 0.1, and uncertainty to 1.5. For Cutie, Livos, and XMem, we apply different memory configurations to accommodate videos of varying lengths. For sequences longer than 200 frames, we use max\_mem\_frames=45, min\_mem\_frames=40, and topk=50, whereas for shorter sequences with fewer than 200 frames, we use max\_mem\_frames=15, min\_mem\_frames=14, and topk=40, thereby enhancing the model's tracking and segmentation capabilities in long-video scenarios. Additionally, to further enhance the robustness of predictions, we employ a Test-Time Augmentation (TTA) strategy, which fuses predictions from both the original and horizontally flipped frames to generate the final masks.  

\textbf{Phase 2: Multi-Model Fusion.}  
To integrate the strengths of different models across various scenarios, we generate final results by aggregating the outputs of different models. Specifically, pixel-level foreground decisions are determined by aggregating confidence scores across models, such that a pixel is classified as foreground if its cumulative score exceeds a predefined threshold. At the object level, a voting mechanism resolves prediction inconsistencies of object IDs among multiple models, producing globally consistent results. Through this approach, we can effectively address challenges arising from overlapping objects and mitigate target disappearance in long video sequences.
  
\label{sec:formatting}

\subsection{2nd Team in VOS Track}

\teaminfo{Transsion}{An Yan, Leilei Cao, Feng Lu, Ran Hong, Fengjie Zhu, Youhai Jiang}{{TEX AI, Transsion Holdings}}

\begin{figure}[ht]
  \centering
  \includegraphics[width=\linewidth]{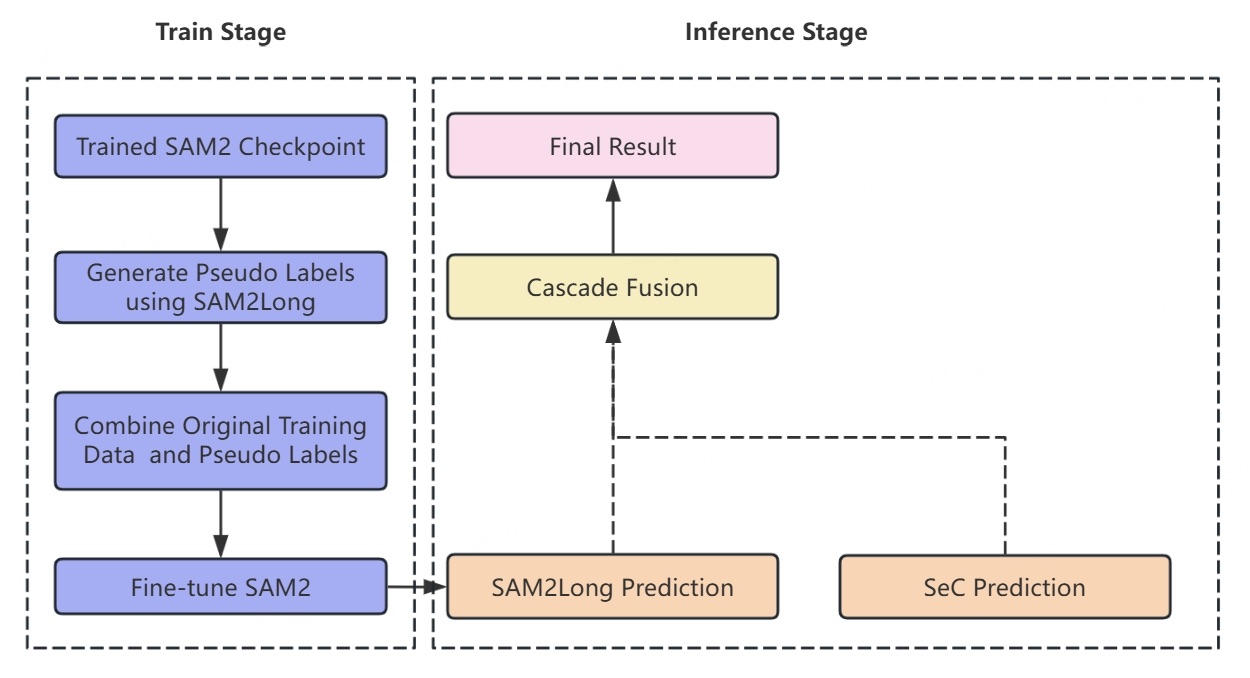} 
  \caption{Overview of our method: pseudo-label enhanced SAM2Long training and cascaded inference with SeC.}
  \label{fig:flowchart}
\end{figure}

Our solution for the LSVOS 2025 VOS Track is built upon the SAM2 framework, enhanced by pseudo-label based domain adaptation and cascaded inference with the SeC model. Fig.~\ref{fig:flowchart} illustrates the overall pipeline, which consists of four main components: (1) a SAM2Long-based baseline, (2) pseudo-label generation, (3) retraining with pseudo labels, and (4) cascaded multi-model inference.

\textbf{Baseline: SAM2 and SAM2Long. }
We adopt the Segment Anything Model 2 (SAM2) as our core segmentation backbone due to its strong generalization capability across diverse video object segmentation scenarios. Specifically, we employ the ViT-L variant of SAM2 as the base architecture. To better handle long video sequences, we leverage the SAM2Long framework, which extends SAM2's temporal modeling capacity via memory propagation mechanisms, making it well-suited for datasets such as MOSE.
In our implementation, the ViT-L variant of SAM2 is first fine-tuned on the target-domain data. The resulting checkpoint serves as the foundation for subsequent enhancements.

\textbf{Pseudo-label Generation. }
To bridge the domain gap between the available training data and the MOSE test distribution, we adopt a pseudo-labeling strategy. Specifically, the fine-tuned SAM2 (ViT-L) checkpoint is used within the SAM2Long framework to generate segmentation masks for each frame of the unlabeled MOSE test set. No additional post-processing or low-confidence filtering is applied, ensuring the complete output distribution is preserved. These generated pseudo labels are then combined with the original MOSE training data to produce an augmented dataset.

\textbf{Retraining with Pseudo Labels. }
The augmented dataset (original training data + MOSE pseudo-labeled test data) is used to further fine-tune the SAM2 model. All data augmentation strategies from the official SAM2 fine-tuning protocol are retained, including random scaling, random cropping, horizontal flipping, and color jittering. The model is trained with a batch size of 1 for 45 epochs, following the same optimizer, learning rate schedule, and loss functions as in the original SAM2 fine-tuning setup. This retraining significantly improves the model's robustness in the MOSE test domain.

\subsection{3rd Team in VOS Track}

\teaminfo{TS\_Video}{Yujie Xie$^{1}$, Hongyang Zhang$^{1,2}$, Zhihui Liu$^{1}$, Shihai Ruan$^{1}$}{{$^{1}$Truesight Research} {$^{2}$The Chinese University of Hong Kong, Shenzhen}}

Given a video with $T$ frames $\{I_t\}_{t=1}^T$, the ground-truth mask $M_1$ of the target objects are provided in the first frame. The goal is to predict segmentation masks $\{M_t\}_{t=2}^T$ for the remaining frames through the segmentation model $f_{\theta}(\cdot)$.

\noindent\textbf{Image Encoder.} 
We adopt Hiera \cite{r3}, a hierarchical masked autoencoder, as the image encoder. Its multiscale architecture enables effective capture of both local details and long-range dependencies, providing robust representations for video segmentation. 

\noindent\textbf{Mask Encoder.} 
The mask encoder in SAM2 encodes segmentation masks by first embedding the input mask through a convolutional module, which projects it into the feature space. This embedding is then element-wise combined with the corresponding frame features from the image encoder, followed by lightweight convolutional layers for feature fusion. 
During tracking, only initialization masks or predicted masks are used, while interactive inputs such as clicks or bounding boxes are excluded to ensure full automation. This design refines mask representations in a compact and efficient manner, enabling precise segmentation and seamless integration into the overall SAM2 pipeline.

\noindent\textbf{Memory Bank.} 
The memory bank stores the initialization frame with its ground-truth mask and the six most recent frames with predicted masks. 
Temporal encodings are applied to recent frames to preserve ordering, while the initialization frame remains unencoded to serve as a target prior. 

\noindent\textbf{Mask Decoder.} 
Current-frame features attend to memory frames to obtain memory-conditioned representations, which are decoded into three candidate masks with IoU scores. 
The mask with the highest score is selected as output, and the memory is updated in a first-in-first-out manner, with the initialization frame permanently retained.

\noindent\textbf{Optimization.}  
The training objective of the proposed method combines complementary losses for pixel-level accuracy, region alignment, overlap quality, and mask-score regression. Concretely, we use a binary cross-entropy (BCE) loss for pixel-wise foreground/background classification, an IoU loss for region-level alignment, a Dice loss to mitigate class imbalance, and a Mask loss to supervise the decoder's predicted mask quality scores.

\begin{equation}
\mathcal{L} = \lambda_{1}\mathcal{L}_{\text{BCE}} + 
              \lambda_{2}\mathcal{L}_{\text{IoU}} + 
              \lambda_{3}\mathcal{L}_{\text{Dice}} +
              \lambda_{4}\mathcal{L}_{\text{Mask}},
\end{equation}

\noindent where the $\mathcal{L}_{\text{Mask}}$ term is defined as:
\begin{equation}
\mathcal{L}_{\text{Mask}} = \frac{1}{K}\sum_{k=1}^{K} \ell\!\big(\hat{s}_k,\, s_k\big), \quad
s_k = \mathrm{IoU}(\hat{M}_k, M_{\text{gt}}),
\end{equation}
with $\hat{s}_k$ the decoder's predicted IoU for candidate mask $\hat{M}_k$, $s_k$ the ground-truth IoU computed against $M_{\text{gt}}$, $K$ the number of candidates per frame, and $\ell(\cdot,\cdot)$ a regression loss (e.g. Smooth-$L_1$ or MSE). The weights $\lambda_{1..4}$ balance the terms.

\begin{figure}[t!]
\centering
\includegraphics[width=2.8in,height=3.8in]{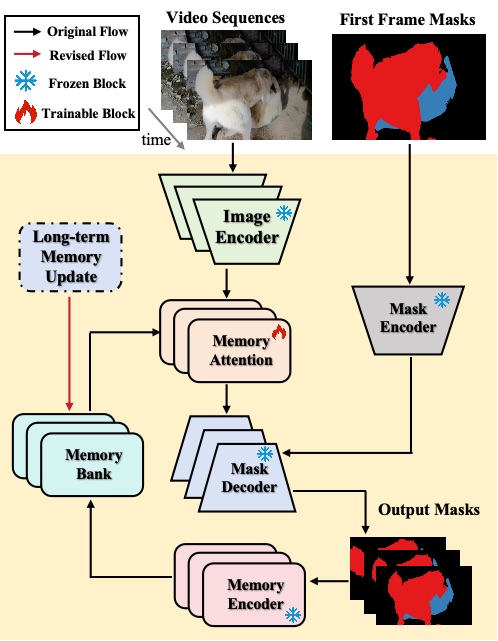}
\caption{Overview of the proposed second stage of training pipeline, where only the memory attention module is fine-tuned during this process.}
\label{fig:train}
\end{figure}

Since the SeC framework adaptively balances LVLM-based semantic reasoning with feature matching and dynamically allocates computation according to scene complexity, and given its superior empirical performance over state-of-the-art methods such as SAM2 and its variants across multiple benchmarks, we adopt it as our baseline. The training framework for the second stage is illustrated in Figure.\ref{fig:train}.

\begin{figure*}[ht!]
	\centering
	\includegraphics[width=\textwidth]{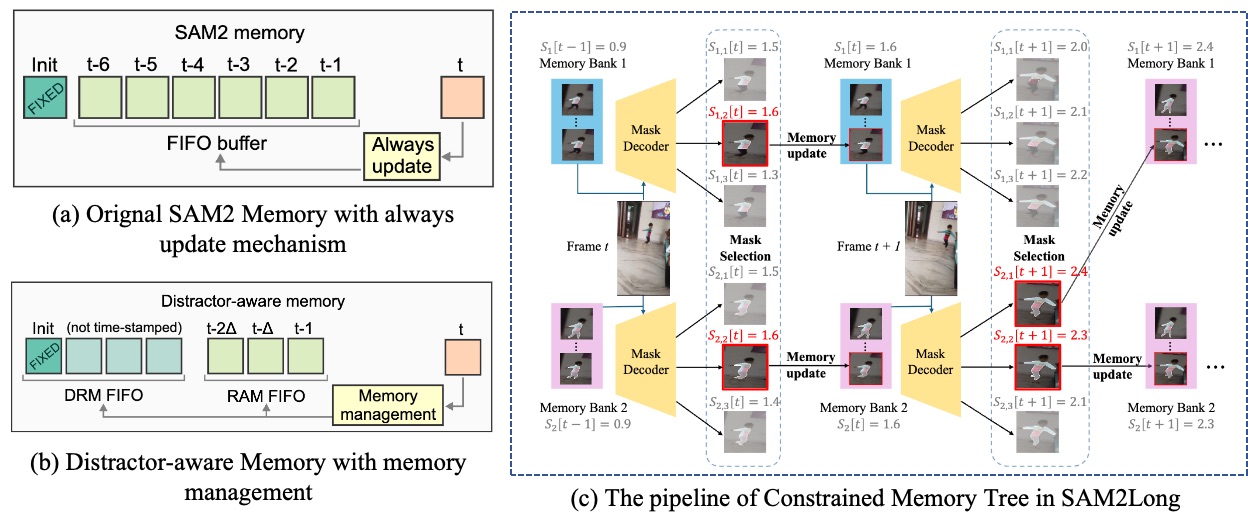}
	\caption{(a) Original design of the SAM2 memory mechanism and (b) proposed distractor-aware memory mechanism, both presented in \cite{videnovic2025distractor}; (c) At each time step, multiple memory pathways are maintained, with the mask decoder generating candidate masks conditioned on memory banks. The pathway with the highest cumulative score is selected for propagation, adapted from \cite{ding2024sam2long}.}
	\label{fig:overview}
\end{figure*}

\noindent\textbf{Long-Term Memory Update for SAM. }
We utilize the grounding encoder from SeC model and enhance the memory bank update mechanism by incorporating a distractor-aware memory module, drawing inspiration from DAM4SAM, to improve robustness and accuracy in video object segmentation. In the inference stage, SAM2Long~\cite{ding2024sam2long} is adopted for robust long-term video object segmentation, using a training-free memory tree to mitigate error accumulation and enable accurate tracking across extended sequences with occlusions.

\noindent\textbf{Sec Model}
Inspired by the Segment Concept (SeC) framework, we adopt its progressive concept-grounding encoder to construct high-level, object-centric representations for video object segmentation. SeC model is trained by the strategy as below:

\noindent\textbf{Concept Guidance with LVLM.}  
To strengthen concept-level reasoning, a sparse keyframe bank is maintained and updated during tracking. 
It retains the initialization frame and a few representative keyframes to ensure semantic diversity. 
LVLM encodes this compact set, with a special \texttt{<SEG>} token extracting object-level concept guidance. 

\noindent\textbf{Scene-Adaptive Activation.}  
To avoid redundancy, a scene-adaptive strategy applies concept guidance only when notable scene changes occur; otherwise, lightweight pixel-level matching is used. 
When activated, the LVLM-derived concept vector is fused with current frame features via cross-attention, enriching memory-enhanced representations. 
This balances semantic priors with fine-grained visual cues, ensuring robust segmentation across challenging scenarios.

\noindent\textbf{Distractor-Aware Memory Strategy. }
To address long-term dependencies, we scale up the memory module inspired by the design of a distractor-aware memory (DAM). Figure~\ref{fig:overview} illustrates the memory management mechanisms in video object segmentation. Figure.\ref{fig:overview} (a) depicts the original SAM2 memory system with an always-update mechanism, featuring a FIFO buffer that processes frames from $t=6$ to $t=1$, with the initial frame fixed and the most recent frame always updated. Figure.\ref{fig:overview} (b) shows the distractor-aware memory management, incorporating a DFM FIFO for fixed initial frames (non-time-stamped), an RAM FIFO for dynamic frames from $t=2$ to $t=1$, and an integrated memory management module for enhanced robustness against distractors. The target of it mainly focused on tracking design, our memory stores an expanded set of temporal features \( \{F_t, C_t\}_{t=1}^T \), with a capacity increased by a factor of \( k \) (e.g., \( k=5 \)) relative to DAM. This larger memory retains detailed object information, critical for handling reappearances after prolonged occlusions. The distractor-aware mechanism computes a similarity score to filter irrelevant objects:
\begin{equation}
S_t = \text{Sim}(C_t, M_t), \quad M_t = \{F_i, C_i \mid i \in [1, t-1]\},
\end{equation}
where \( \text{Sim}(\cdot,\cdot) \) is a cosine similarity function, and \( M_t \) is the memory bank. Low-scoring distractors are suppressed, ensuring focus on the target object. The distractor-aware mechanism, adapted from DAM4SAM, enhances accuracy by mitigating interference from similar objects.

\noindent\textbf{SAM2Long for inference. }
During inference, we further introduce SAM2Long to improve robustness without introducing additional training costs. The method adopts a constrained tree memory structure with uncertainty handling. 

The detailed information of constrained tree memory is illustrated in Figure.\ref{fig:overview} (c). Formally, given a set of memory nodes \(\{m_i\}_{i=1}^N\), each associated with an uncertainty score \(\sigma_i\), the aggregated memory feature at time step \(t\) is computed as:
\begin{equation}
\hat{M}_t = \sum_{i=1}^{N} w_i \cdot m_i, \quad 
w_i = \frac{\exp\left(-\sigma_i\right)}{\sum_{j=1}^N \exp\left(-\sigma_j\right)},
\end{equation}
where the weights \(w_i\) are constrained by the tree hierarchy, ensuring that closer parent-child nodes in the memory tree receive consistent weighting. 
The uncertainty score \(\sigma_i\) is estimated from prediction confidence, allowing unreliable memory nodes to be down-weighted automatically. 

The ensemble mechanism then fuses the uncertainty-aware memory \(\hat{M}_t\) with the concept representation \(C_t\), yielding the final segmentation prediction:
\begin{equation}
\hat{Y}_t = f_{\text{dec}}(\hat{M}_t, C_t, I_t),
\end{equation}
where \(I_t\) denotes the current frame embedding and \(f_{\text{dec}}\) is the mask decoder. 

This design not only balances adaptability and stability, but also mitigates error accumulation by dynamically suppressing noisy or outdated memory entries. 
Consequently, SAM2Long achieves consistent tracking under long-term occlusion, re-appearance, and large-scale appearance variations, while maintaining high efficiency at test time.

\section{RVOS Track Top Solution}
\label{sec:mevis_method}
The top three teams of the RVOS track are reported in Table~\ref{tab:results_rvos}. The first place team achieved a $\mathcal{J}\&\mathcal{{F}}$ score of 67.33\% on the testing set.

\begin{table}[!ht]
    \renewcommand\arraystretch{1.1}
    \centering
    \setlength\tabcolsep{10pt}
    \caption{Top 3 winners of the RVOS Track.}
    \vspace{-3mm}
    \small
    {\begin{tabular}{rlccc}
            \toprule
             Rank & Team  & {$\mathcal{J}$} & {$\mathcal{{F}}$} & {$\mathcal{J\&{F}}$} \\
             \midrule
             \textcolor{gold}{\faTrophy} 1 & \textbf{SaSaSa2VA}   & \textbf{63.82} & \textbf{70.84} & \textbf{67.33} \\
             \textcolor{silver}{\faMedal} 2 & Transsion    & 61.29 & 68.01 & 64.65 \\
             \textcolor{bronze}{\faMedal} 3 & dytino & 61.06 & 67.22 & 64.14 \\
            \bottomrule
        \end{tabular}}%
    \label{tab:results_rvos}%
\end{table}%

\subsection{1st Team in RVOS Track}

\teaminfo{SaSaSa2VA}{Quanzhu Niu$^{1}$, Dengxian Gong$^{1}$, Shihao Chen$^{1}$, Tao Zhang$^{1}$, Yikang Zhou$^{1}$, Haobo Yuan$^{2}$, Lu Qi$^{1}$, Xiangtai Li$^{3}$, Shunping Ji$^{1}$}{{$^{1}$Wuhan University}~{$^{2}$University of California, Merced}~{$^{3}$Nanyang Technological University}}

\begin{figure*}[t]
  \centering
  \includegraphics[width=0.97\linewidth]{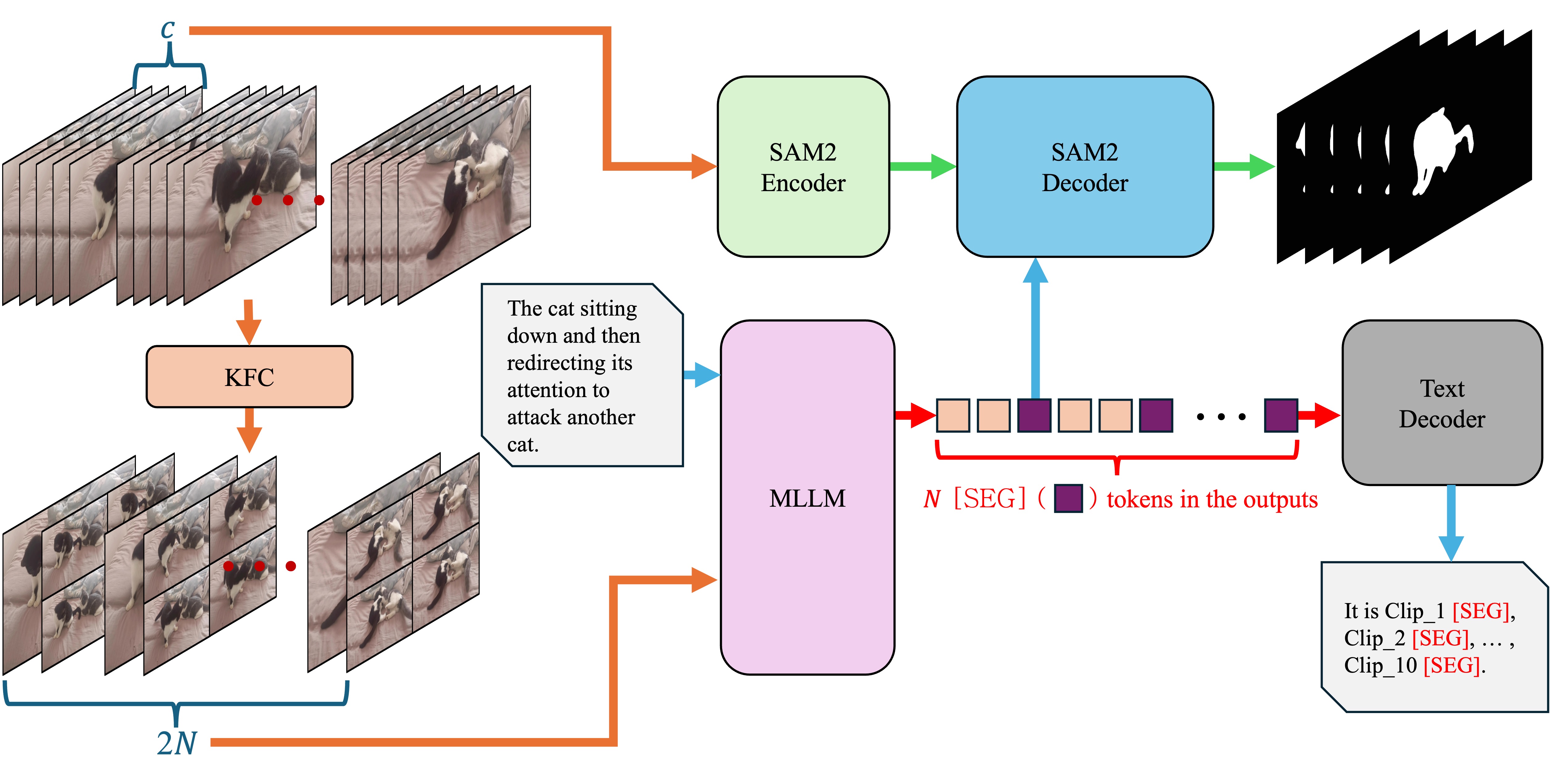}
  \caption{\textbf{Overview of Segmentation Augmentation in SaSaSa2VA. }We design Key Frame Compression (KFC) strategy and Scaling~\texttt{[SEG]} tokens strategy over Sa2VA~\cite{yuan2025sa2va}. A $T$-frame video is divided into $N$ non-overlapping clips, each containing $c = g^2{+}1$ frames. The frames passed to the MLLM are then compressed via KFC. The MLLM outputs $N$~\texttt{[SEG]} tokens, each corresponding to one clip. For a given clip, conditioned on the original $c$ frames and the hidden state of its~\texttt{[SEG]} token, SAM2 decodes the masks for that clip. In this figure, $c$ is set to $5$, resulting $g=2$.  }
  \label{fig:sa}
\end{figure*}

\textbf{Baseline: Sa2VA}
\label{sec:method_sa2va}
\noindent\textbf{Meta Architecture. }Sa2VA~\cite{yuan2025sa2va} comprises an Multi-modal Large Language Model (MLLM)~\cite{chen2024expanding} and SAM2~\cite{ravi2024sam}. The MLLM accepts images, videos, and text instructions as input and produces text responses conditioned on the instructions. When the user instruction requests segmentation results, the text response includes the segmentation token~\texttt{[SEG]}. The hidden state of the segmentation token serves as an implicit prompt, which SAM2 converts into object segmentation masks at both the image and video levels.
Sa2VA adopts InternVL 2.5~\cite{chen2024expanding}. InternVL 2.5 follows a LLaVA-like~\cite{liu2023visual} architecture composed of an InternVIT~\cite{chen2024internvl}, an MLP projector, and a Large Language Model (LLM)~\cite{cai2024internlm2technicalreport, qwen2025qwen25technicalreport}. Images and videos are encoded by InternViT~\cite{chen2024internvl} into visual tokens, which are projected by an MLP and combined with text tokens as input to the LLM. The LLM autoregressively generates text responses that may include~\texttt{[SEG]} tokens. The hidden state of the~\texttt{[SEG]} token from the last LLM transformer layer is processed by an MLP to form the prompt input to SAM2~\cite{ravi2024sam}.

\noindent\textbf{SAM2. }SAM2~\cite{ravi2024sam} produces object segmentation masks for selected high-resolution video frames based on the segmentation prompts from the MLLM. It then propagates these frame-level masks to obtain object segmentation results for the entire video.

\label{sec:method_sa}
\noindent\textbf{Limitations of Sa2VA. }To reduce time and memory consumption, only five frames are sampled per video during Sa2VA~\cite{yuan2025sa2va} training, and each video uses a single~\texttt{[SEG]} token to transmit information between the MLLM and SAM2. During inference, the MLLM processes only five frames for video understanding, and a single~\texttt{[SEG]} token is then used to propagate masks across the entire video. Sampling only five frames inevitably limits the MLLM's ability to capture global video context, and relying on a single~\texttt{[SEG]} token to convey segmentation information for the whole video struggles to accommodate temporal changes in object position, shape, and even appearance/disappearance. Consequently, this design imposes limitations on segmentation performance.

\noindent\textbf{Overview. }As illustrated in~\cref{fig:sa}, we design several Segmentation Augmentation strategies over Sa2VA~\cite{yuan2025sa2va}, including Key Frame Compression (KFC) and Scaling~\texttt{[SEG]} tokens. Details are described below.

\noindent\textbf{Key Frame Compression. }To balance spatiotemporal efficiency with the MLLM's global understanding of videos, we propose a Key Frame Compression (KFC) scheme. We sample $T = N\times c$ frames from the original video and denote the sequence as $V=\{I_1, I_2,\dots, I_{T}\}$, where each frame $I_t\in \mathbb{R}^{H\times W \times 3}$ is an RGB image at time step $t$. We then divide the sequence into $N$ non-overlapping clips, each containing $c = g^2+1$ frames. Each clip is denoted as $C^{i} = \{I^{i}_1, I^{i}_2,\dots, I^{i}_{c} \}$ for $i= 1, 2, \dots, N$. In each clip $C^{i}$, the first frame $I^{i}_1$ is the key frame, and the remaining $c-1=g^2$ frames $\{I^{i}_2, I^{i}_3,\dots, I^{i}_{c} \}$ are compressed. Specifically, we tile these $g^2$ frames into a $g\times g$ grid image $I^i_{cat}\in\mathbb{R}^{gH\times gW \times 3}$ in row-major order (left to right, top to bottom), and resize the grid back to $H\times W$ to obtain the compressed image $I^i_{com}\in\mathbb{R}^{H\times W \times 3}$:
\begin{equation}
    I^i_{cat} = \mathrm{concatenate}(I^{i}_2, I^{i}_2,\dots, I^{i}_{c}),
\end{equation}
\begin{equation}
    I^i_{com} = \mathrm{resize}(I^i_{cat}).
\end{equation}

In this way, each clip $C^{i}$ sends only one key frame and one compressed image $\{I^{i}_1, I^i_{com}\}$ to the MLLM, reducing a video with $T$ frames to just $2N$ images. This approach preserves global video information while mitigating redundant attention to adjacent frames.
\noindent\textbf{Scaling~\texttt{[SEG]} tokens. }To handle diverse temporal variations of the objects, we increase the number of~\texttt{[SEG]} tokens. Specifically, we assign one~\texttt{[SEG]} token to each clip $C^{i}$, so the MLLM produces $N$~\texttt{[SEG]} tokens per video. The hidden states of these tokens are denoted by $S=\{s^1, s^2, \dots , s^N\}, s^i \in \mathbb{R}^d$, where $d$ is the hidden dimension. In SAM2, $s^i$ is used to decode the masks ($M^i=\{m^i_1, m^i_2, \dots, m^i_{c}\}, m^i_j \in \{0,1\}^{H\times W}$) of the object within clip $C^i$. The process is described by:
\begin{equation}
    S = \mathrm{MLLM}(I^1_1, I^1_{com}, I^2_1, I^2_{com}, \dots, I^N_1, I^N_{com}),
\end{equation}
\begin{equation}
    m^i_j = \mathrm{SAM2}(I^i_j, s^i).
\end{equation}

Specifically, during training, we supervise only the mask $m^i_1$ of the key frame $I^i_1$ in each clip.

\noindent\textbf{Inference sampling strategies. }During training, we sample $T = N\times c$ frames per video using a specific procedure, whereas at inference we must accommodate videos of varying lengths. To this end, we design five sampling strategies, each exhibiting advantages for different videos.

\begin{itemize}
    \item \textbf{Uniform. }Regardless of video length, the video is evenly divided into $N$ ori-clips. In each ori-clip, uniformly sample $c$ frames as the clip sent to the MLLM. In SAM2, the frames corresponding to each ori-clip need to be decoded using the associated $s$.
    \item \textbf{Uniform+. }Building on the Uniform strategy, for videos whose original length is shorter than $T$ frames, some frames near clip boundaries have 2 corresponding~\texttt{[SEG]} tokens. We average the masks from the 2 results.
    \item \textbf{Q-frame. }We use the method in~\cite{qframe} to select the top $T$ frames most related to the text prompt. The selected frames are then sorted in temporal order and processed with Uniform+.
    \item \textbf{Wrap-around. }Given a target of $T$ frames, sample cyclically using indices $i\bmod T_{ori}$, where $T_{ori}$ is the original video length. If $T_{ori}\geq T$, this yields the first $T$ frames; the masks beyond $T$ are propagated by SAM2's memory. If $T_{ori}< T$, it wraps around and repeats until $T$ frames are collected, and the selected frames are then sorted in temporal order.
    \item \textbf{Wrap-around+. }When $T_{ori}< T$, we use Wrap-around strategy. When $T_{ori}\geq T$, we instead use the Uniform strategy.
\end{itemize}

\noindent\textbf{Selective Averaging. }Different inference sampling strategies perform differently across videos, and models of different scales also differ in their video understanding, leading to variations in final scores~\cite{fang20251stsolution4thpvuw}. To leverage their complementary strengths, we adopt a Selective Averaging scheme. For each mask, the results from different models and sampling methods are weighted and averaged. If the weighted average value of a pixel exceeds 0.5, its mask value is set to 1; otherwise, it is set to 0.

\subsection{2nd Team in RVOS Track}

\teaminfo{Transsion}{Ran Hong$^{1,2}$, Feng Lu$^{1,3}$, Leilei Cao$^{1}$, An Yan$^{1}$}{{$^{1}$TEX AI, Transsion Holdings} {$^{2}$Nanchang University} {$^{3}$ShanghaiTech University} }

Consider a video sequence consisting of $T$ frames $\mathcal{V} = \{\mathbf{I}_t\}_{t=1}^{T}$, where each frame $\mathbf{I}_t \in \mathbb{R}^{3 \times H \times W}$ represents an RGB image of height $H$ and width $W$. Given a referring expression $\mathcal{T} = \{w_i\}_{i=1}^{L}$, where $w_i$ denotes the $i$-th token, the goal of RVOS is to predict a sequence of binary masks $\mathcal{M} = \{\mathbf{M}_t\}_{t=1}^{T}$ with $\mathbf{M}_t \in \{0,1\}^{H \times W}$, such that $\mathbf{M}_t$ identifies the spatial region of the object referred to by $\mathcal{T}$ in frame $\mathbf{I}_t$. Formally, RVOS can be defined as learning a mapping $f: (\mathcal{V}, \mathcal{T}) \mapsto \mathcal{M}$.

\begin{figure*}[t] %
    \centering
    \includegraphics[width=\textwidth]{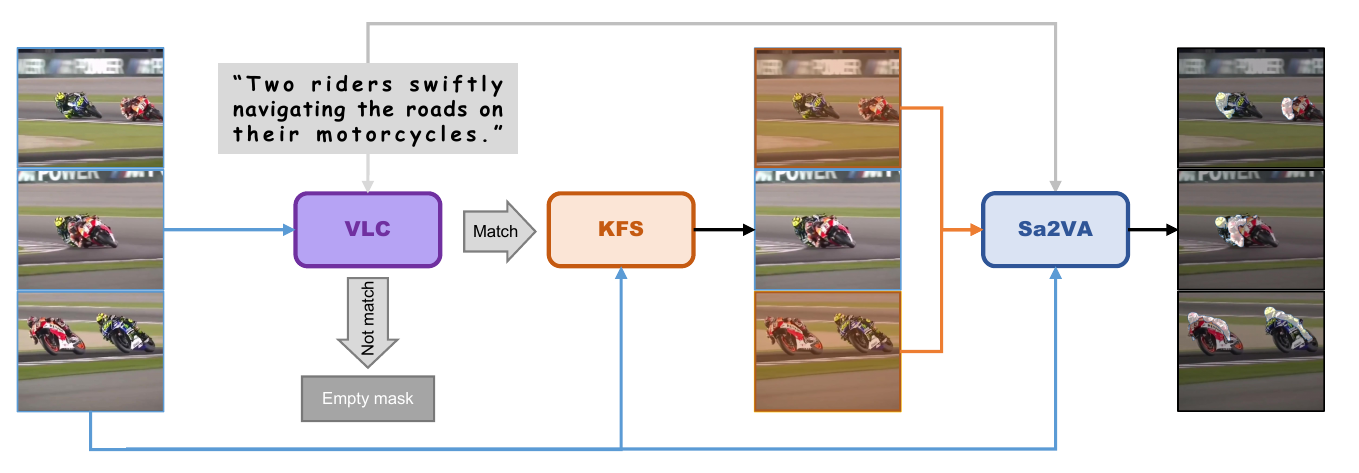}
    \caption{Overall framework of our method.}
    \label{fig:overall}
\end{figure*}

As shown in the Figure \ref{fig:overall}, Our proposed approach for RVOS is composed of three sequential modules.

\textbf{VLC} evaluates the correspondence between the input video and the referring expression, ensuring that subsequent segmentation is performed only when the text matches the video content.

\textbf{KFS} selects a set of informative key frames from the video to reduce temporal redundancy and provide focused input for the segmentation module.

\textbf{Sa2VA} takes the sampled key frames, the preceding video frames, and the referring expression as input, and sequentially predicts the binary masks of the referred object in each frame. This modular design enables our framework to first verify textual-visual relevance, focus on informative frames, and finally generate temporally consistent segmentation masks aligned with the referring expression.

\textbf{Video Semantic Matching}
The RVOS task can be understood as segmenting the object in a video that matches a given textual description. This matching involves both subject-level correspondence and action-level correspondence. If the video does not contain the subject described in the text, or if the subject's action does not correspond to the described action, a zero mask should be output.

To address this, we employ a large pre-trained model from QwenVL \cite{Qwen-VL} to perform video-text correspondence verification. Specifically, in the VLC module, the video sequence and the corresponding referring expression are input to the QwenVL along with the prompt:

"Please check whether the video matches the input text, i.e., whether the subject described in the text exists in the video and whether the subject's action corresponds to the action described in the text. Output yes/no."

If the model outputs no, all segmentation masks are set to zero. Otherwise, the video proceeds to subsequent modules for key-frame sampling and object segmentation.

\textbf{Key Frame Sampler.}
Compared to conventional video object segmentation (VOS) tasks, the RVOS task poses additional challenges due to the need for semantic understanding and precise alignment between the textual description and video content.

The original Sa2VA model is a multi-task framework not specifically designed for RVOS. In its default configuration, Sa2VA selects the first five frames as key frames. However, for datasets such as MeVis, the object corresponding to the textual description may not appear in the first few frames, and action understanding often requires observing more frames. In Sa2VA, the key frames and the referring expression are input into a large pre-trained model to learn a SEG token, which is then used to segment the object in the key frames. Since the output is always a single SEG token regardless of the number of key frames, selecting too many frames can reduce the expressive capacity of the SEG token. Therefore, a reasonable key-frame selection strategy is critical to improving segmentation accuracy.

A straightforward approach is uniform sampling across the video, effectively compressing the original video. However, uniform sampling suffers from several limitations: the optimal number of sampled frames is difficult to determine, and for long videos, consecutive sampled frames may be too far apart, resulting in a loss of action information. Observing that most objects corresponding to the text appear in the early part of the video, we adopt a hybrid strategy combining head continuous sampling with uniform sampling. This approach controls the number of key frames to maintain the SEG token's expressive capacity while also capturing action dynamics in longer videos, leading to more accurate segmentation of the referred objects.

\textbf{Sa2VA for segmentation.}
After sampling key frames and obtaining the SEG token from the LLM in Sa2VA, the SEG token is used as a prompt for the SAM2 Decoder to segment the object in the key frames. Subsequently, the masks obtained from the key frames are propagated as prompts through SAM2 across the entire video, enabling the segmentation of all frames corresponding to the textual description. This two-stage process-first segmenting the key frames and then propagating the masks-ensures that the model captures both the object appearance and its temporal dynamics, yielding accurate segmentation for all frames in the video.

The entire procedure of our framework can be summarized as follows: for each frame $t$, the segmentation mask $\mathbf{M}_t$ is given by

\[
\mathbf{M}_t =
\begin{cases}
\mathbf{0}, & \text{if } f_\text{VLC}(\mathcal{V}, \mathcal{T}) = 0, \\
f_\text{Sa2VA}(\mathbf{I}_1, \dots, \mathbf{I}_t, \{\mathbf{I}_{t_k}\}_{k=1}^{N}, \mathcal{T}) , & \text{if } f_\text{VLC}(\mathcal{V}, \mathcal{T}) = 1
\end{cases}
\]

where $f_\text{VLC}$ denotes the VLC module that determines whether the video $\mathcal{V}$ contains the subject and action described in the referring expression $\mathcal{T}$. The set $\{\mathbf{I}_{t_k}\}_{k=1}^{N}$ represents the $N$ key frames sampled from the video. The function $f_\text{Sa2VA}$ represents the Sa2VA model, which first segments the key frames using the SEG token as a prompt for the SAM2 decoder and then propagates the masks across the entire video. In this way, each mask $\mathbf{M}_t$ depends only on the first $t$ frames, the sampled key frames, and the referring expression.

\subsection{3rd Team in RVOS Track}

\teaminfo{dytino}{Alexey Nekrasov$^{1}$, Ali Athar, Daan de Geus$^{2}$, Alexander Hermans$^{1}$, Bastian Leibe$^{1}$}{{$^{1}$RWTH Aachen University} {$^{2}$Eindhoven University of Technology}}

\begin{figure}[t]
    \centering
    \includegraphics[width=\linewidth]{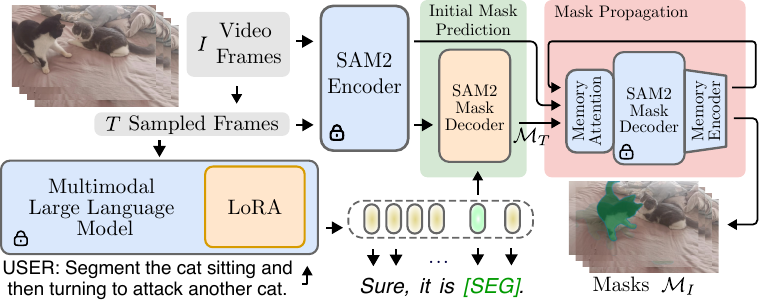}
    \caption{
    \textbf{Sa2VA-i Architecture Overview.}
    Sa2VA-i first predicts initial masks $\mathcal{M}_T$ using a \textit{finetuned} SAM2 mask decoder by taking the generated \texttt{[SEG]} token and predicting a mask for all $T$ sampled frames separately. Next, it propagates these  $\mathcal{M}_T$ masks across all video frames $I$ using SAM2's \textit{original} mask decoder, yielding output masks $\mathcal{M}_I$.
    }
    \label{fig:mainfig}
\end{figure}

Our model, called Sa2VA-i and shown in \cref{fig:mainfig}, addresses this problem by ensuring the inference procedure is consistent with the training procedure.
To achieve this, during inference, we do \textit{not} use SAM2's memory components to predict segmentation masks and follow the exact same procedure that is followed during training.

Concretely, we take the predicted \texttt{[SEG]} token's features and the per-frame video features and feed them directly to the mask decoder, to predict segmentation masks $\mathcal{M}_T$ for the $T$ sampled frames.
Subsequently, to make a prediction for all $I$ frames of the video, we use an off-the-shelf, \textit{non-finetuned} SAM2 decoder weights.
Concretely, we directly prompt SAM2 with predicted masks $\mathcal{M}_T$, and use the original SAM2 inference procedure to predict masks $\mathcal{M}_I$ for all frames.
By following this approach, there is no longer any incompatibility between non-finetuned memory components and a finetuned mask decoder, because (a) no memory components are used to make the initial predictions $\mathcal{M}_T$, and (b) the SAM2 components used to obtain the final masks $\mathcal{M}_I$ are all original and thus compatible.

In practice, this means that we have to store weights for two versions of the mask decoder: (a) the original one from SAM2, and (b) the finetuned one from Sa2VA.
The other components from SAM2 remain frozen when training Sa2VA, so the same weights can be used for initial mask prediction with Sa2VA-i and mask propagation with SAM2.
This means that there is only a small additional memory footprint of $\sim$16MB.

Additionally, we observe that the original Sa2VA uses random frame sampling during training, but samples the first $T$ video frames during inference.
This is suboptimal due to the offline nature of RVOS -- with prompts like ``the dog that disappears from the left, then re-appears'' -- where full videos have to be available during inference to answer the question properly.
Therefore, we propose to apply uniform frame sampling during inference instead.
Furthermore, to ensure further consistency between training and inference, we also train a version of Sa2VA-i that applies uniform frame sampling during training as well.
We find that these simple improvements significantly improve performance.

\section{Conclusion and Discussion}
\label{sec:conclusion}

In this year's LSVOS challenge, we have seen record diversity of participants, from both industry and academia worldwide. The enthusiasm and depth of participation underscore the growing importance of Video Object Segmentation. Several takeaways stand out from the top solutions.
First, the newly introduced MOSEv2 track reveals substantial remaining headroom for modern VOS methods. The numerical comparisons of the results with MOSEv1 show that MOSEv2 is markedly more difficult, with notable drops in leading methods' scores, evidence that current state-of-the-art systems still struggle in complex, realistic scenarios.
Second, large language models (LLMs) have effectively become a default component in many pipelines, especially for language-guided video tasks, highlighting their promise for video understanding. We anticipate that deeper integrations of LLMs will continue to lift performance.
Looking ahead, we will focus on the most challenging failure modes identified by this year's results and by real-world use cases, with the aim of further advancing the frontier of Video Object Segmentation and related research.

{
    \small
    \bibliographystyle{ieeenat_fullname}
    \bibliography{main}
}

\end{document}